\documentclass[runningheads]{llncs}
\usepackage{graphicx}
\usepackage{amsmath,amssymb} % define this before the line numbering.
\usepackage{color}
\usepackage{multirow}
\usepackage{wrapfig}
\usepackage{color}
\usepackage[width=122mm,left=12mm,paperwidth=146mm,height=193mm,top=12mm,paperheight=217mm]{geometry}
\begin{document}
% \linenumbers
\pagestyle{headings}
\mainmatter

\title{Local Binary Pattern Networks} % Replace with your title

\titlerunning{Local Binary Pattern Networks}

\authorrunning{Jeng-Hau Lin, Yunfan Yang, Rajesh Gupta, Zhuowen Tu}

\author{Jeng-Hau Lin$^1$, Yunfan Yang$^1$, Rajesh Gupta$^1$, Zhuowen Tu$^{2,1}$}

%Please write out author names in full in the paper, i.e. full given and family names. 
%If any authors have names that can be parsed into FirstName LastName in multiple ways, please include the correct parsing, in a comment to the volume editors:
%\index{Lastnames, Firstnames}
%(Do not uncomment it, because you may introduce extra index items if you do that...)

\institute{$^1$Department of  Computer Science and Engineering, UC San Diego\\
	$^2$Department of  Cognitive Science, UC San Diego\\
	\email{\{jel252,yuy130,rgupta,ztu\}@ucsd.edu}
}

\maketitle

\begin{abstract}
% 1. Save memory/costjeng hau 
% 2. Possible speedup
% 3. Backprop closed form
% 4. Experimental results on MNIST, SVHN, and CIFAR-10

Memory and computation efficient deep learning architectures are crucial to continued proliferation of machine learning capabilities to new platforms and systems. Binarization of operations in convolutional neural networks has shown promising results in reducing model size and computing efficiency. 
In this paper, we tackle the problem using a strategy different from the existing literature by proposing local binary pattern networks or LBPNet, that is able to learn and perform binary operations in an end-to-end fashion. LBPNet\footnote{U.S. Provisional Application Attorney Docket No.: 009062-8375.US00/SD2018-044-1} uses local binary comparisons and random projection in place of conventional convolution (or approximation of convolution) operations. These operations can be implemented efficiently on different platforms including direct hardware implementation. We applied LBPNet and its variants on standard benchmarks. The results are promising across benchmarks while providing an important means to improve memory and speed efficiency that is particularly suited for small footprint devices and hardware accelerators.

        
\end{abstract}

%%%%%%%%% BODY TEXT
\vspace{-4mm}
\section{Introduction}
\vspace{-2mm}
% 1. CNN overburdens IoT
% 2. Different mainstream technologies in CV, and we promote LBP
% 3. An overview of LBP net
% 	3.1 Primitive binary operation
%	3.2 deformable
%	3.3 non-convolution
%	3.4 end-to-end
% 3. Contributions
%	3.1 less memory
%	3.2 no multiplier, no adder
%	3.3 mathematic closed form
%Since 1989 LeCun proposed convolutional neural network~\cite{lecun1989backpropagation} (CNN), CNNs has been the dominant technique in supervised learning particularly for image classification.
%Later on, numerous powerful neural network models have been proposed, such as AlexNet~\cite{krizhevsky2012imagenet}, VGG-16~\cite{simonyan2014very} and inception~\cite{szegedy2017inception}.
Convolutional Neural Networks (CNN) \cite{lecun1989backpropagation} have had a notable impact on many applications. Modern CNN architectures such as AlexNet \cite{krizhevsky2012imagenet}, VGG \cite{simonyan2015very}, GoogLetNet \cite{szegedy2015going}, and ResNet \cite{he2015resnet} have greatly advanced the use of deep learning techniques \cite{Hinton06} into a wide range of computer vision applications \cite{girshick2014rich,long2015fully}. These gains have surely benefited from the continuing advances in computing and storage capabilities of modern computing machines.
% With CNN-family models being increasingly popular in applications and real products, practicality issues regarding the model size and computation speed also become important factors to consider.
Table~\ref{tab:history} lists recognition accuracy, number of parameters, model size, and floating point operations (FLOP), for three well-known architectures \cite{krizhevsky2012imagenet,simonyan2015very,szegedy2015going}. While there have been improvements, these model sizes and computational demands primarily require the use of desktop- or server-class machines in real-world applications.

\begin{table}[!htp]
\vspace{-3mm}
\small
\centering
\caption{\footnotesize Model size and computational requirements for three well-known CNN architectures for the classification on ImageNet.}
\vspace{-3mm}
\scalebox{0.8}{
\begin{tabular}{l|lll} 
                & AlexNet \cite{krizhevsky2012imagenet}    & VGG16 \cite{simonyan2015very}      & GoogLeNet \cite{szegedy2015going}  \\ \hline
Accuracy        & 84.7\%      & 92.38\%     & 93.33\%     \\
Parameters       & 61 million  & 138 million & 6.8 million \\
Memory          & 233MB       & 526MB       & 26MB        \\
FLOP            & 1.5 billion & 3 billion   & 1.5 billion
\end{tabular}
}
\vspace{-3mm}
\label{tab:history}
\end{table}
As CNN-family models mature and take on increasingly complex pattern recognition tasks, the commensurate increase in the use of computational resources further limits their use to compute-heavy CPU and GPU platforms with sophisticated (and expensive) memory systems. By contrast, the emerging universe of embedded devices especially when used as edge-devices in distributed systems presents a much greater range of potential applications. These systems can enable new system-level services that use sophisticated in-situ learning and analysis tasks. The primary obstacle to this 
vision is the need for significant improvements in memory and computational efficiency of deep learning networks both in their model size as well as working set size. 

Various methods have been proposed to perform network pruning \cite{lecun1989optimal,guo2016dynamic}, compression \cite{han2015deep,iandola2016squeezenet}, or sparsification\cite{liu2015sparse}. Impressive results have been achieved lately by using binarization of selected operations in CNNs \cite{courbariaux2015binaryconnect,hubara2016binarized,rastegari2016xnornet}. At the core, these efforts seek to approximate the internal computations from floating point to binary while keeping the underlying convolution operation exact or approximate. 
% However, these approaches trying to binarize CNNs: (1) are still limited to convolution-based operations and (2) leave a gap between binarization and float-point based convolution.

\begin{figure}[!htp]
\vspace{-4mm}
	\centering
	\includegraphics[width=0.6\columnwidth]{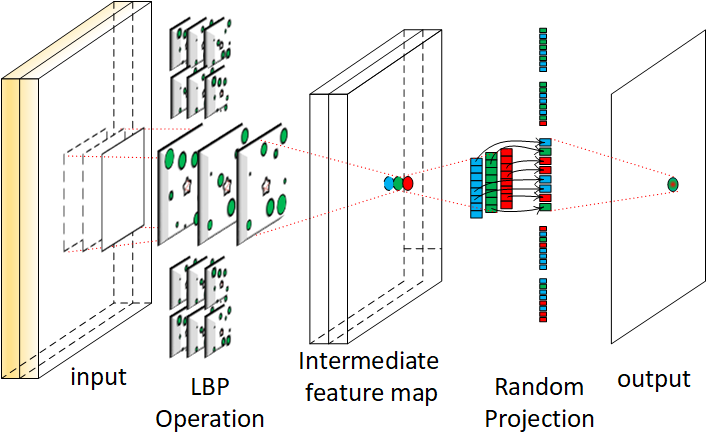}
    \vspace{-4mm}
	\caption{\footnotesize An illustration of the LBPNet architecture. The LBP operation is composed of comparison and bit-allocation, while the channel fusion is done by random projection.}
    \label{fig:overview}
    \vspace{-5mm}
\end{figure}

In this paper, we explore an alternative using non-convolutional operations that can be executed in an architectural and hardware-friendly manner, trained in an end-to-end fashion from scratch (distinct to the previous attempts of binarizing the CNN operations). We note that this work has roots in research before the current generation of deep learning methods. Namely, the adoption of 
%Jenghau, SIFT was not used here. removing "two types of features, spatial invariant feature transform (SIFT) \cite{lowe2004distinctive} and" 
local binary patterns (LBP) \cite{ojala1996comparative}, which uses a number of {\em predefined} sampling points that are mostly on the perimeter of a circle, to compare with the pixel value at the center. The combination of multiple logic outputs (``$1$" if the value on a sampling point is greater than that on the center point and ``$0$" otherwise) gives rise to a surprising rich representation \cite{wang2009hog} about the underlying image patterns and has shown to be complementary to the SIFT-kind features \cite{lowe2004distinctive}. However, LBP has been under-explored in the deep learning research community where the feature learning part in the existing deep learning models \cite{krizhevsky2012imagenet,he2015resnet} primarily refers to the CNN features in a hierarchy. 
%(not the neural network architecture though) are learned automatically from the data.
In the past, despite some existing attempts \cite{juefei2016local}, the logic operation (comparison) in LBP was not directly made into the existing CNN frameworks, due to the intrinsic difference between the convolution and comparison operations.

%Our approach to local binary pattern networks (LBPNet) mainly builds upon the local binary pattern operators \cite{ojala1996comparative}.
Here, we aim at building a hardware friendly neural network architecture by learning and executing binary operations in an end-to-end fashion. We name our algorithm {\em local binary pattern networks} (LBPNet). 
We note that LBPNet performs non-convolutional comparisons instead of arithmetic operations.
All the binary logic operations in LPBNet are directly learned, which is in a stark distinction to previous attempts that try to either binarize CNN operations \cite{hubara2016binarized,rastegari2016xnornet} or to approximate LBP with convolution operations \cite{juefei2016local}.
%Unlike the recent work by \cite{juefei2016local} where convolution is used to approximate the LBP operation, we directly learn the logic operations from scratch in an end-to-end fashion.
In the current CNN frameworks, operations like max-pooling and ReLU can be made logical since no addition or multiplication operations are needed. This makes the convolution and fusion (implemented mostly by summation of channels or $1\times 1$ convolution) to be the main computational challenge. We solve it by deriving a differentiable function to learn the binary pattern and adopt random projection for the fusion operations. Fig.~\ref{fig:overview} illustrates the overview of LBPNet. The resulting LBPNet can thus be trained end-to-end using binary operations. Results show that thus configured LBPnet achieves modest results on benchmark datasets but it is a significant improvement in the parameter size reduction gain (hundreds) and speed improvement (thousand times faster).
%We also show an ablation to systematically replace the standard convolutional layer in CNN by the LBPNet layer.
Our experiments demonstrate the value of LBPNet in embedded system platforms to enable emerging internet of things (IoT) applications.

\vspace{-2mm}
\section{Related Works}
\vspace{-2mm}
\label{sec:related}
% 1. Hubara's BNN
% 2. XNOR-Net
% 3. Deformable CNN
% 4. LBCNN
Related work falls along three primary dimensions. 

\noindent{\bf Binarization for CNN}. Binarizing CNNs to reduce the model size has been an active research direction \cite{courbariaux2015binaryconnect,hubara2016binarized,rastegari2016xnornet}.
The main focus of \cite{courbariaux2015binaryconnect} is to build binary connections between the neurons. The binarized neural networks work (BNN) \cite{hubara2016binarized} successfully broke the curse of dimensionality as it relates to precision in hardware.
Through binarizing both weights and activations, the model size was reduced, and the multiplication can be replaced by logic operation. Non-binary operations like batch normalization with scaling and shifting are still implemented in floating-point~\cite{hubara2016binarized}.
%However, the binarization was an extremely aggressive quantization and resulted in an inferior accuracy. 
%Hubara et. al. relied batch normalization with scaling and shifting to bring back the accuracy of the BNN.
As a result, BNN is not totally bit-wise but it intelligently moves the inter-neuron traffic to intra-neuron computation.
%Although BNN succeeded on MNIST, CIFAR-10, and SVHN, it suffered too much on ImageNet dataset.
The XNOR-Net~\cite{rastegari2016xnornet} introduces extra scaling layer to compensate the loss of binarization, and achieves a state-of-the-art accuracy on ImageNet. Both BNNs and XNORs can be considered as the discretization of real-numbered CNNs, while the core of the two works are still based on spatial convolution.

\noindent{\bf CNN approximation for LBP operation}. Recent work on local binary convolutional neural networks (LBCNN) in \cite{juefei2016local} takes an opposite direction to BNN \cite{hubara2016binarized}. LBCNN utilizes subtraction between pixel values together with a ReLU layer to simulate the LBP operations.
The convolution between the sparse binary filters and images is actually a difference filtering, thus making LBCNN work like an edge detector.
%Therefore, they applied LBCNN on residual networks structure to bring back the low frequency information.
During the training, the sparse binarized difference filters are fixed, only the successive $1$-by-$1$ convolution serving as channel fusion mechanism and the parameters in batch normalization layers are learned.
However, the feature maps of LBCNN are still in floating-point numbers, resulting in significantly increased model complexity as shown in Tables \ref{tab:mnist} and \ref{tab:cifar10}.
By contrast, LBPNet learns binary patterns and logic operations from scratch, resulting in orders of magnitude reduction in memory size and testing speed up than LBCNN.

\noindent {\bf Active or deformable convolution}. Among notable line of recent work that learns local patterns are active convolution \cite{jeon2017active} and deformable convolution \cite{dai2017deformable}, where data dependent convolution kernels are learned. Both of these are quite different from LBPNet since they do not seek to improve the network efficiency.
%The related aspect is in learning control points for the templates where the leaning of offset field is done with a truncated difference filter \cite{dai2017deformable}.
%The offset kernels enforce the second set of convolutional kernels to learn on image gradients and collect the vector sums to push the positions of the elements in the first set of convolutional kernels. 
Our binary patterns learn the position of the sampling points in an end-to-end fashion as logic operations that do not have the addition operations whereas \cite{dai2017deformable} essentially learns data-dependent convolutions. 
\vspace{-2mm}
\section{Local Binary Pattern Network}
\vspace{-2mm}
An overview of the LBPNet architecture is shown in Fig.~\ref{fig:overview}.
The forward propagation is composed of two steps: LBP operation and channel fusion.
We introduce the patterns in LBPNets and the two steps in the following sub-sections, and then move on to the engineered network structures for LBPNets.
%which can be achieved by bit-wise random projection or $1$-by-$1$ convolution. LBCNN~\cite{juefei2016local} resorted to $1$-by-$1$ convolution to learn a linear combination of channels. Instead, to avoid multiplication-and-accumulation operations, we use a fixed random projection for the channel fusion of the intermediate future maps. 

\vspace{-2mm}
\subsection{Patterns in LBPNets}
\label{sec:forward}
% 1. Design Choice
% 2. Indexing and Comparison
% 3. Some details in implementation:
%	3.1 bit shifting
%	3.2 tensor manipulation

\begin{figure}[ht]
    \vspace{-8mm}
	\centering
	\includegraphics[width=0.7\columnwidth]{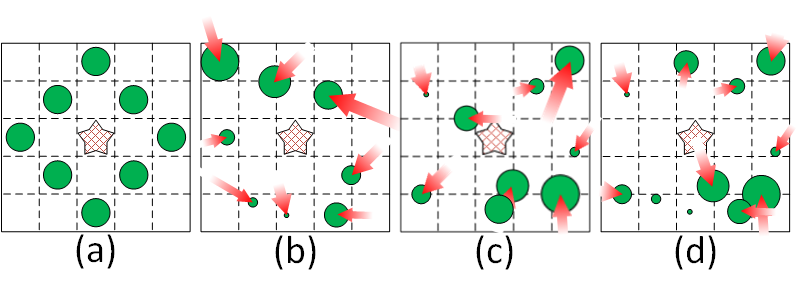}    
    \vspace{-4mm}
	\caption{\footnotesize (a) A traditional local binary pattern is a fixed pattern without much variety. (b)-(d) In this paper, local binary patterns are initialized using a normal distribution of positions within a given area, followed by end-to-end supervised learning. The red arrows denote pushing forces during training.}
    \label{fig:pattern1}
    \vspace{-4mm}
\end{figure}

In LBPNet, there are multiple patterns defining the positions of sampling points to generate multiple output channels.
Patterns are randomly initialized with a normal distribution of locations on a predefined square area, and then subsequently learned in an end-to-end supervised learning fashion.
Fig.~\ref{fig:pattern1} (a) shows a traditional local binary pattern; there are eight sampling points denoted by green circles, surrounding a pivot point in the meshed star at the center of pattern; Fig.~\ref{fig:pattern1}(b)-(d) shows a learnable pattern with eight sampling points in green, and a pivot point as a star at the center.
Different sizes of the green circle stand for the bit position of the true-false outcome on the output magnitude.
We allocate the comparison outcome of the largest green circle to the most significant bit of the output pixel, the second largest to the 2nd bit, and so on.
The red arrows represents the driving forces that can push the sampling points to better positions to minimize the classification error.
%The 4 sampling points are confined in a 5x5 area.
We describe the details of forward propagation in the following two sub-sections.

\subsection{LBP Operation}
\vspace{-2mm}

\begin{figure}
	\centering
    \vspace{-4mm}
    \begin{tabular}{c}
	\includegraphics[width=0.8\columnwidth]{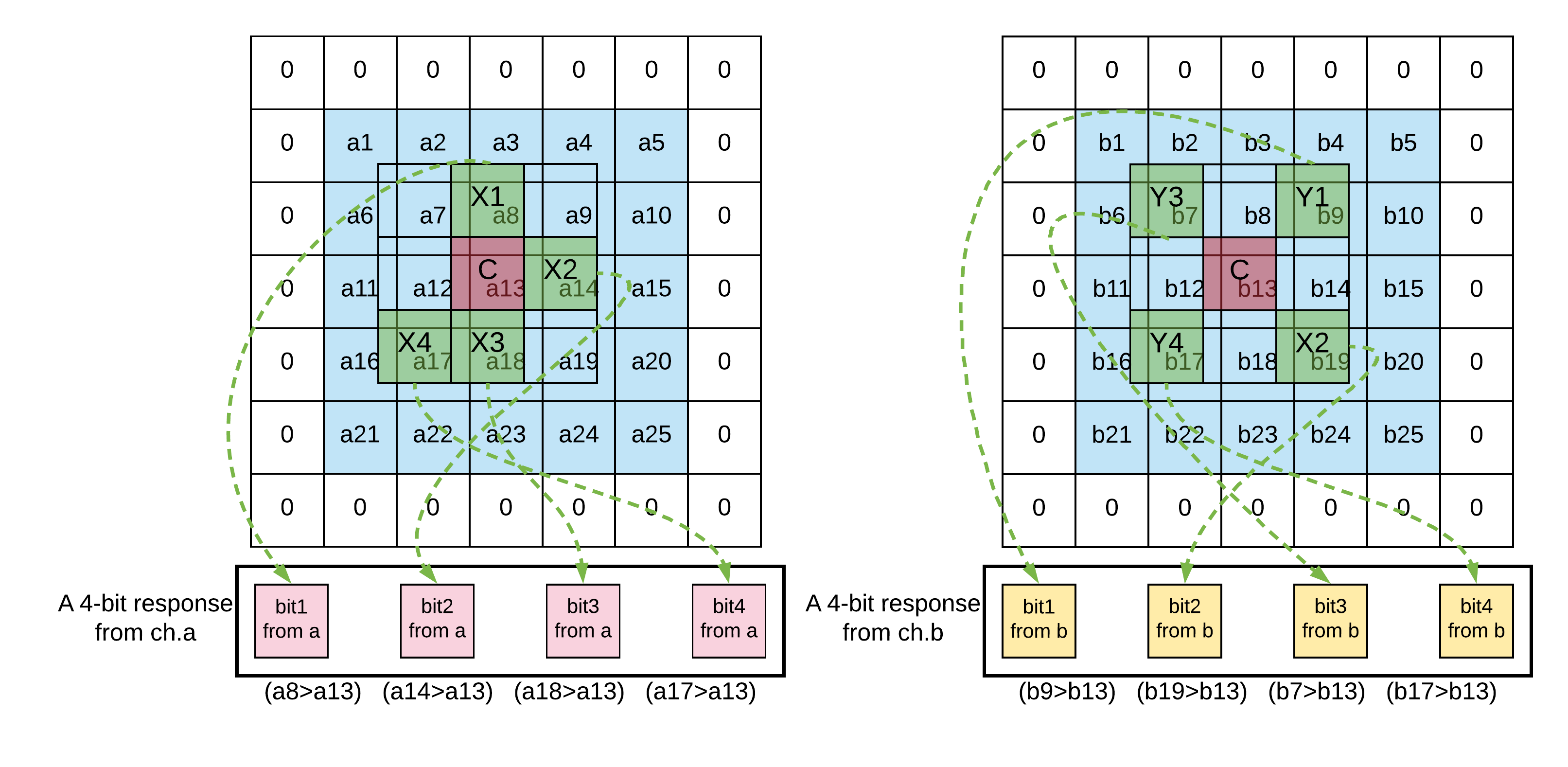}
    \end{tabular}
    	\vspace{-4mm}
    \begin{tabular}{cc}
    (a) LBP operations for channel ch.a; & (b) LBP operations for channel ch.b.
    \end{tabular}
	\vspace{+2mm}
	\caption{An example of LBP operation on multiple input channels. Each local binary pattern has 4 sampling points restricted within $3$-by-$3$ areas.}
    \label{fig:LBP_inP2}
  \vspace{-4mm}
\end{figure}

First, LBPNet samples pixels from incoming images and compares the sampled pixel value with the center sampled point, the pivot.
If the sampled pixel value is larger than that of the center one, the output is a bit ``$1$"; otherwise, the output is set to ``$0$."
Next, we allocate the output bits to a number's different binary digits based on a predefined ordering.
The number of sampling points defines the number of bits of an output pixel on a feature map.
Then we slide the local binary pattern to the next location and perform the aforementioned steps until a feature map is generated.
In most case, the incoming image has multiple channels, hence we perform the aforementioned comparison on every input channel.

Fig.~\ref{fig:LBP_inP2} shows a snapshot of the LBP operations.
Given two input channels, ch.a and ch.b, we perform the LBP operation on each channel with different kernel patterns.
The two 4-bit response binary numbers of the intermediate output are shown on the bottom.
For clarity, we use green dashed arrows to mark where the pixel are sampled, and list the comparison equations under each bit.
A logical problem has emerged: we need a channel fusion mechanism to avoid the explosion of the channels.

\subsection{Channel Fusion with Random Projection}

\begin{figure}[t]
	\centering
	\includegraphics[width=0.9\columnwidth]{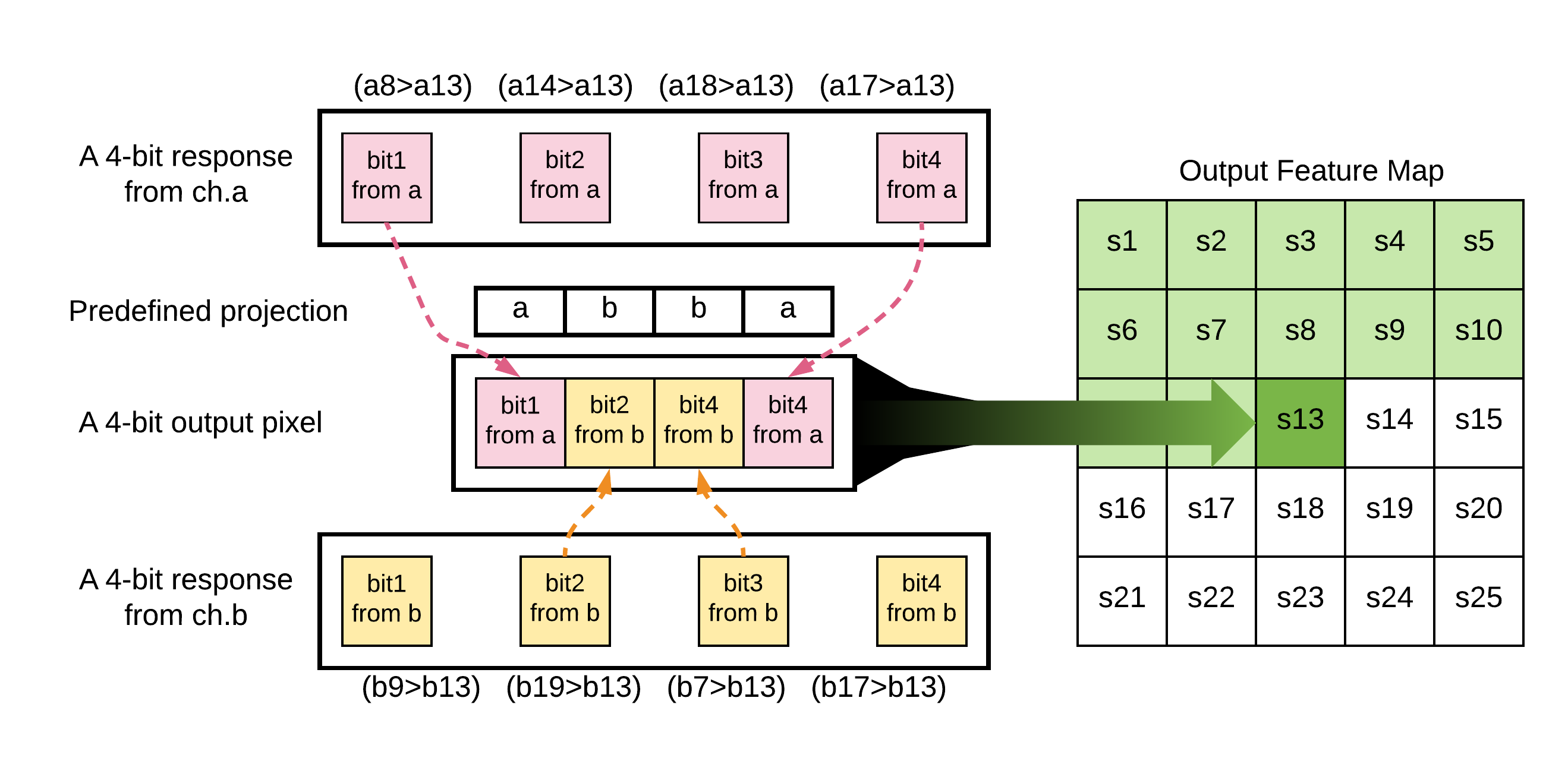}
    \vspace{-4mm}
	\caption{An example of LBP channel fusing. The two 4-bit responses from Fig.~\ref{fig:LBP_inP2} are fused together and assigned to pixel ``$s13$'' on the output feature map.}
    \label{fig:LBP_in2_rndPrj}
    \vspace{-4mm}
\end{figure}

We use random projection~\cite{bingham2001random} as a dimension-reducing and distance-preserving process to select output bits among intermediate channels for the concerned output channel as shown in Fig.~\ref{fig:LBP_in2_rndPrj}. 
The random projection is implemented with a predefined mapping table for each output channel.
The projection map is fixed upon initialization.
All output pixels on the same output channel share the same mapping.
In fact, random projection not only solves the channel fusion with a bit-wise operation, but also simplifies the computation, because we do not have to compare all sampling points with the pivots.
For example, in Fig.~\ref{fig:LBP_in2_rndPrj}, the two pink arrows from intermediate ch.a, and the two yellow arrows from intermediate ch.b bring the four bits for the composition of an output pixel.
Only the MSB and LSB on ch.a and the middle two bits on ch.b need to be computed.
If the output pixel is $n$-bit, for each output pixel, there will be $n$ comparisons needed, which is irrelevant with the number of input channels.
The more input channels simply bring the more combinations of channels in a random projection table.

Throughout the forward propagation, there is no resource demanding multiplication or addition. 
Only comparison and memory access are used. 
Therefore, the design of LBPNets is efficient in the aspects of both software and hardware.

\vspace{-3mm}
\subsection{Network structures for LBPNet}
\label{sec:structure}
\vspace{-3mm}

\begin{figure}[ht]
	\vspace{-6mm}
	\centering
	\includegraphics[width=0.75\columnwidth]{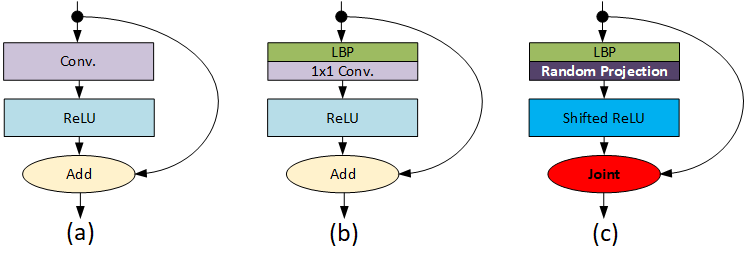}
    \vspace{-4mm}
	\caption{\footnotesize Basic LBPNet blocks. (a) the well-known building block of residual networks. (b) The transition-type building block, in which a $1$-by-$1$ convolutional layer is used for the channel fusion of a preceding LBP layer. (c) The multiplication and accumulation (MAC) free building block for LBPNet.}
    \label{fig:ResNet-like}
    \vspace{-5mm}
\end{figure}
The network structure of LBPNet must be carefully designed.
Owing to the nature of comparison, the outcome of a LBP layer is very similar to the outlines in the input image.
In other words, our LBP layer is good at extracting high frequency components in the spatial domain, but relatively weak at understanding low frequency components.
Therefore, we use a residual-like structure to compensate this weakness of LBPNet.

Fig.~\ref{fig:ResNet-like} shows three kinds of residual-net-like building blocks. 
Fig.~\ref{fig:ResNet-like} (a) is the typical building block for residual networks.
The convolutional kernels learn to obtain the residual of the output after the addition.
Our first attempt is to introduce the LBP layer into this block as shown in Fig.~\ref{fig:ResNet-like} (b), in which we utilize a $1$-by-$1$ convolution to learn a better combination of LBP feature maps.
However, the convolution incurs too many multiplication and accumulation operations especially when the LBP kernels increases.
Then, we combine LBP operation with a random projection as mentioned in previous sections.
Because the pixels in the LBP output feature maps are always positive, we use a shifted rectified linear layer (shifted-ReLU) to increase nonlinearities.
The shifted-ReLU truncates any magnitudes below the half the the maximum of the LBP output. 
More specifically, if a pattern has $n$ sampling points, the shifted-ReLU is defined as Eq.~\ref{eq:shifted_ReLU}. 
\begin{equation}
f(x) = \left\{\begin{matrix}
x           & \text{, } x>  2^{n-1}-1\\ 
2^{n-1}-1   & \text{, otherwise}
\end{matrix}\right.
\label{eq:shifted_ReLU}
\vspace{-2mm}
\end{equation}
As mentioned earlier, the low-frequency components can be lost when the information is passing through several LBP layers.
To make the block totally MAC-free, we use a joint operation to cascade the input tensor of the block and the output tensor of the shifted-ReLU along the channel dimension.
Although the jointing of tensors brings back the risk of channel explosion, the number of channels can be controlled if we carefully design the number of LBP kernels.

\vspace{-3mm}
\subsection{Hardware Benefits}
\label{sec:hardware}
\vspace{-2mm}

\begin{wraptable}{r}{0.5\linewidth} \small
\vspace{-12mm}
\small
\centering
\caption{The number of logic gates for arithmetic units. Energy use data for technology node: $45$nm.}
\vspace{-0mm}
\scalebox{0.9}{
\begin{tabular}{||cc|rr||}
\hline
Device Name           & \#bits  & \#gates  & Energy (J) \\ \hline
\multirow{2}{*}{Adder}&  4      &  20      & $\le$ 3E-14  \\
					  & 32      & 160      & 9E-13      \\ \hline
Multiplier            & 32      & $\ge$144 & 3.7E-12    \\ \hline
Comparator            &  4      & 11       & $\le$ 3E-14  \\
\hline
\end{tabular}
}
\label{tab:hardware}
\vspace{-4mm}
\end{wraptable}

LBPNet saves in hardware cost by avoiding the convolution operations. 
Table~\ref{tab:hardware} lists the reference numbers of logic gates of the concerned arithmetic units.
A ripple-carry full-adder requires $5$ gates for each bit.
A 32-bit multiplier includes a data-path logic and a control logic.
Because there are too many feasible implementations of the control logic circuits, we conservatively use an open range to express the sense of the hardware expense.
The comparison can be made with pure combinational logic circuit of $11$ gates, which also means only the infinitesimal internal gate delays dominate the computation latency.
Comparison is not only cheap in terms of its gate count but also fast due to a lack of sequential logic inside.
Slight difference on numbers of logic gates may apply if different synthesis tools or manufacturers are chosen.
%Obviously, the saving of hardware cost is more than $10$ times if we change a multiplier to a comparator.
Assuming the capability of a LBP layer is as strong as a convolutional layer in terms of classification accuracies.
Replacing the convolution operations with comparison directly gives us a $27$X saving of hardware cost.

Another important benefit is energy saving.
The energy demand for each arithmetic device has been shown in~\cite{horowitz20141}.
If we replace all convolution operations with comparisons, the energy consumption is reduced by $153$X. 

\section{Backward Propagation of LBPNet}
\label{sec:backward}
\vspace{-2mm}
% 1. Turn non-differentiable to differentiable
% 2. Use image gradient to push and learn better patterns
To train LBPNets with gradient-based optimization methods, we need to tackle two problems:
1). The non-differentiability of comparison; and
2). the lack of a source force to push the sampling points in a pattern.

\subsection{Differentiability}
\begin{figure}[!htp]
%\begin{wrapfigure}{r}{0.5\linewidth} \small
	\vspace{-7mm}
	\centering
	\includegraphics[width=0.5\columnwidth]{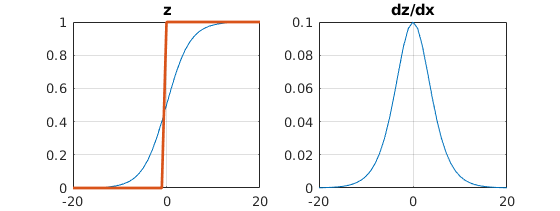}
    \vspace{-3mm}
	\caption{We approximate the comparison operation with a scaled and shifted hyperbolic tangent function. On the left sub-figure, the orange line denotes the comparison function, and the blue curve is the differentiable approximation. The right sub-figure shows the derivative of the approximation for convex optimization. The scaling parameter is $10$ in this figure.}
    \label{fig:tanhApprox}
    \vspace{-2mm}
%\end{wrapfigure}
\end{figure}

The first problem can be solved if we approximate the comparison operation shown in Eq.~\ref{eq:comparison} with a shifted and scaled hyperbolic tangent function as shown in Eq.~\ref{eq:tanhApprox}.
\vspace{-2mm}
\begin{equation}
I_p>I_c
\label{eq:comparison}
\end{equation}
\vspace{-2mm}
\begin{equation}
\frac{1}{2}(tanh(\frac{I_p-I_c}{k})+1),
\label{eq:tanhApprox}
\end{equation}
where $k$ is the scaling parameters to accommodate the number of sampling points from a preceding LBP layer. 
The hyperbolic tangent function is differentiable and has a simple closed-form for the implementation, as depicted in Fig.~\ref{fig:tanhApprox}.

\subsection{Deformation with Optical Flow Theory}
To deform the local binary patterns, we resort to the concept from optical flow theory.
Assuming the image content in the same class share the same features, even though there are certain minor shape transformations, chrominance variations or different view angles, the optical flow on these images should share similarities with each others. 
\vspace{-2mm}
\begin{equation}
%\frac{\partial I}{\partial x}\frac{dx}{dt} + \frac{\partial I}{\partial y}\frac{dy}{dt} = -\frac{\partial I}{\partial t}
\frac{\partial I}{\partial x}V_x + \frac{\partial I}{\partial y}V_y = -\frac{\partial I}{\partial t}
\label{eq:opticalFlow}
\vspace{-2mm}
\end{equation}

Eq.~\ref{eq:opticalFlow} shows the optical flow theory, where $I$ is the pixel value, a.k.a luminance, $V_x$ and $V_y$ represent the two orthogonal components of the optical flow among the same or similar image content.
The LHS of optical flow theory can be interpreted as a dot-product of image gradient and optical flow, and this product is the inverse the derivative of luminance versus time across different images.

To minimize the difference between images in the same class is equivalent to extract similar features of image in the same class for classification.
However, both the direction and magnitude of the optical flow underlying the dataset are unknown.
The minimization of a dot-product cannot be done by changing the image gradient to be orthogonal with the optical flow.
Therefore, the only feasible path to minimize the magnitude of the RHS is to minimize the image gradient.
Please note the sampled image gradient $\frac{\partial I}{\partial x}\hat{x}+\frac{\partial I}{\partial y}\hat{y}$ can be changed by deforming the apertures, which are the sampling points of local binary patterns.
%Since the minimizing gradient is the goal of gradient descent, the solution of the optimization problem is the training of LBPNet.

%Based on deduction above, we can simply 
When applying calculus chain rule on the cost of LBPNet with regard to the position of each sampling point, one can easily reach a conclusion that the last term of the chain rule is the image gradient.
Since the sampled pixel value is the same as the pixed value on the image, the gradient of sampled value with regard to the sampling location on a pattern is equivalent to the image gradient on the incoming image.
Eq.~\ref{eq:goal} shows the gradient from the output loss through a fully-connected layer with weights, $w_j$, toward the image gradient.
\vspace{-2mm}
\begin{equation}
\frac{\partial cost}{\partial position} = \sum_{j}(\Delta_j w_j)\frac{\partial g(s)}{
\partial s}\frac{\partial s}{\partial I{i,p}}(\frac{\textbf{d} I_{i,p}}{\textbf{d}x}\hat x + \frac{\textbf{d} I_{i,p}}{\textbf{d}y}\hat y),
\label{eq:goal}
\vspace{-4mm}
\end{equation}
where $\Delta_j$ is the backward propagated error, $\frac{\partial g(s)}{
\partial s}$ is the derivative of activation function, and $\frac{\partial s}{\partial I{i,p}}$ is the gradient of Eq.~\ref{eq:tanhApprox} also plotted in Fig.\ref{fig:tanhApprox}.

\begin{figure}
    \vspace{-4mm}
	\centering
	\includegraphics[width=0.95\columnwidth]{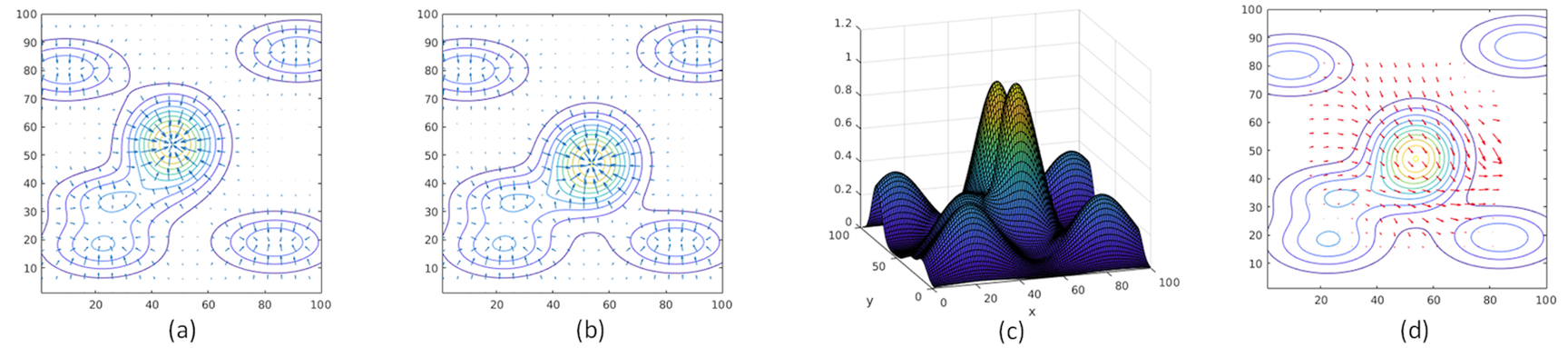}
    \vspace{-4mm}
	\caption{A set of 6 Gaussian peaks, and the center one is moving toward right-bottom slightly. (a) the previous contour map and image gradient; (b) the current contour map and image gradient; (c) a side-view of the Gaussian peaks; (d) the optical flow of the highest Gaussian peak.}
    \label{fig:opticalFlow}
    \vspace{-4mm}
\end{figure}
Fig.~\ref{fig:opticalFlow} illustrates an example of optical flow theory.
In this figure, the highest peak is moving toward the right-bottom, and the image gradients are different.
The calculation of optical flow requires heavy linear algebraic manipulation, and the result is shown in sub-figure (d).
The optical flow reveals the motion of the highest peak and is often used for object tracking.
Utilizing Eq.~\ref{eq:goal} to train LBPNet is to calculate the vector sums over the image gradients.
After the update, the sampling points (apertures) will move downhill or uphill depending on the error $\Delta$ in Eq.~\ref{eq:goal}.
Without computing the optical flow, the sampling points are still pushed to a position with minimal average image gradient and a minimum absolute value of the RHS in Eq.~\ref{eq:opticalFlow} is guaranteed.
%As a result, the difference between images $\frac{\partial I}{\partial t}$, will be minimized.
%Although the image gradients and optical flows of the images in a dataset is constant, if we train the local patterns to adapt local maximums/minimums, the feature maps of hidden layers could be more and more similar to each others.

%In summary, by accumulating the image gradient with Eq.~\ref{eq:goal} as a force to push the sampling points, we can deform the patterns to learn the features of a dataset.

\section{Experiments}
\label{sec:experi}
% 1. Datasets: MNIST, CIFAR-10
% 2. Experimental results
% 3. Discussion: intuition behind the numbers
\vspace{-2mm}

In this section, we conduct a series of experiments on three datasets: MNIST, SVHN, and CIFAR-10 to verify the capability of LBPNet. Here is the description of the datasets and setups.

\subsection{Experiment Setup}

Images in the MNIST dataset are hand-written numbers from $0$ to $9$ in $32$-by-$32$ gray scale bitmap format.
The dataset is composed with a training set of $60,000$ examples and a test set of $10,000$ examples.
The manuscripts were written by both staff and students.
Although most of the images can be easily recognized and classified, there are still a portion of sloppy images inside MNIST. 

SVHN is an image dataset of house numbers.
Although cropped, images in SVHN include some distracting numbers around the labeled number in the middle of the image.
The distracting parts increase the difficulty of classifying the printed numbers.
There are $73,257$ training examples and $26,032$ test examples in SVHN.

CIFAR-10 is composed of daily objects, such as airplanes, cats, dogs, and trucks.
The size of images is in $32$-by-$32$ and has $3$ channels of RGB colors.
The training set includes $50,000$ examples, and the test set includes $10,000$ examples as well.

In all of the experiments, we use all training examples to train LBPNets, and directly validate on test sets.
To avoid peeping, the validation errors are not employed in the backward propagation.
There are no data augmentations used in the experiments. 
Because CIFAR-10 is relatively harder than the other two datasets, we convert RGB channels into YUV channels to improve the classification accuracy.
%\vspace{-5pt}
\begin{figure}[t]
	\centering
	\includegraphics[width=0.75\columnwidth]{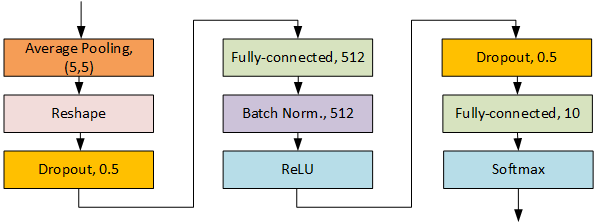}
    \vspace{-4mm}
	\caption{The MLP block that is shared by the variants of our LBPNet models, which is made with two fully-connected layers of $512$ and $10$ neurons. Besides the nonlinearities, there are two Dropout layers with $0.5$ possibility and one batch-normalization layer.}
    \label{fig:classifier}
    \vspace{-10pt}
\end{figure}

The goal of the experiments is to compare LBPNet with convolution-based methods.
We implement two versions of LBPNet using the two building blocks shown in Fig.~\ref{fig:ResNet-like} (b) and (c).
For the remaining parts of this paper, we call the LBPNet using $1$-by-$1$ convolution as the channel fusion mechanism {\bf LBPNet($1$x$1$)} (has convolution in the fusion part), and the version of LBPNet utilizing random projection {\bf LBPNet(RDP)} (totally convolution-free).
The number of sampling points in a pattern is set to $4$, and the area size for the pattern to deform is $5$-by-$5$.
LBPNet also has an additional multilayer perceptron (MLP) block, which is a $2$-layer fully-connected  as shown in Fig.~\ref{fig:classifier}. The MLP block's performance without any convolutional layers or LBP layers on the three datasets is shown in Table~\ref{tab:mnist},~\ref{tab:svhn},~\ref{tab:cifar10}. The model size and speed of the MLP block are excluded in the comparisons since all models have an MLP block.
%that doesn't add much to the overall model complexity and computing time.

\vspace{-2mm}
\subsection{Experimental Results}
\vspace{-2mm}

\begin{table}[!htp]
\vspace{-3mm}
%\begin{wraptable}{r}{0.45\linewidth} \small
%\vspace{-8mm}
\small
\centering 
\caption{The cycle count of different arithmetics involved in the experiments. For the corresponding hardware description, please refer to Sec.~\ref{sec:hardware}.}
\vspace{-2mm}
\scalebox{0.9}{
\begin{tabular}{||l|r||}
\hline
Arithmetic              & \#cycles \\  \hline
32-32bit Multiplication & 4 \\ 
32-1bit Multiplication  & 1 \\
1-1bit Multiplication   & 1 \\
32bit Addition          & 1 \\
4bit Comparison         & 1 \\
\hline
\end{tabular}
}
\label{tab:cycle_count}
\vspace{-2mm}
%\end{wraptable}
\end{table}

To understand the capability of LBPNet when compared with existing convolution based methods, we build two feed-forward streamline CNNs as our baseline.
The basic block of the CNNs contains a spatial convolution layer (Conv) followed by a batch normalization layer (BatchNorm) and a rectified linear layer (ReLU).
For MNIST, the baseline is a 4-layer CNN with kernel number of $40$-$80$-$160$-$320$ before the classifier. 
The baseline CNN for SVHN and CIFAR-10 has 10 layers ($64$-$64$-$128$-$128$-$256$-$256$-$256$-$512$-$512$-$512$) before the classifier because the datasets are larger and include more complicated content.

In addition to the baseline CNNs, we also build three shallow CNNs subject to comparable memory sizes of LBPNets(RDP) to demonstrate the efficiency of LBPNet. We call the shallow CNNs as CNN(lite).
For MNIST, the CNN(lite) model contains only one convolutional layer with $40$ kernels. The CNN(lite) model for SVHN has $2$ convolutional layers ($8$-$17$). 
The CNN(lite) model for CIFAR-10 also has $2$ convolutional layers ($8$-$9$).

In the BNN~\cite{hubara2016binarized} paper, the classification on MNIST is done with a binarized multilayer perceptron network (MLP).
We adopt the binarized convolutional neural network (BCNN) in~\cite{hubara2016binarized} for SVHN to perform the classification and re-produce the same accuracy as shown in~\cite{jeng2017bcnnwsf} on MNIST.

\begin{figure}[!htp]
	\vspace{-4mm}
	\centering
    \includegraphics[width=1.0\columnwidth]{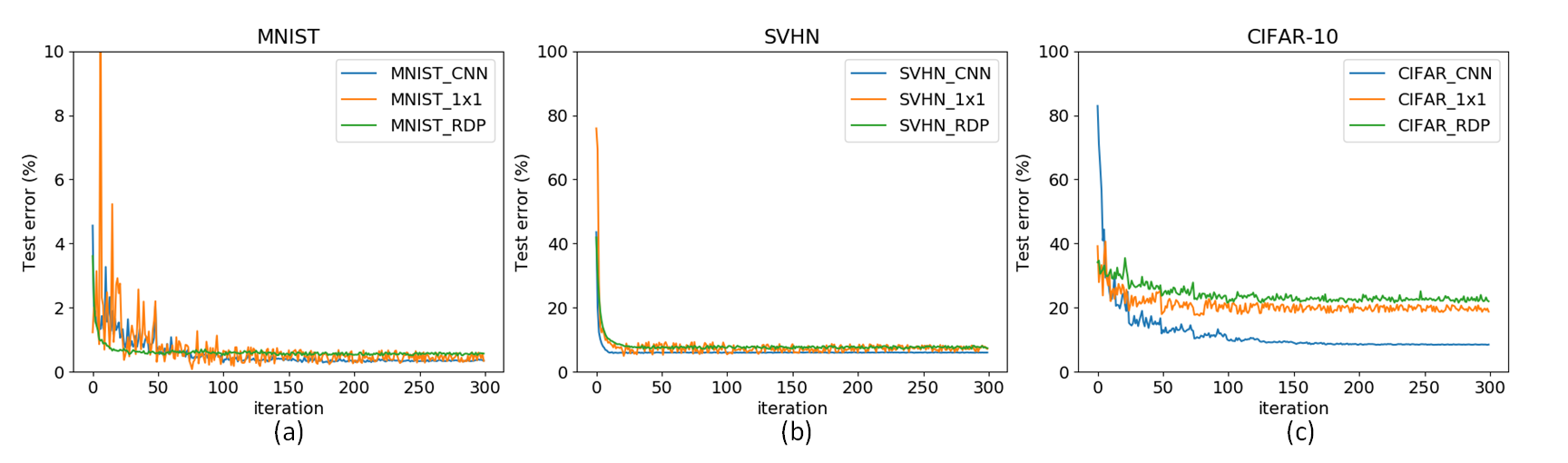}
    \vspace{-10mm}
	\caption{Error curves on benchmark datasets. (a) test errors on MNIST; (b) test errors on SVHN; (c) test errors on CIFAR-10}
    \label{fig:learning_curves}
    \vspace{-6mm}
\end{figure}

The learning curves of LBPNets on the three datasets are plotted in Fig.~\ref{fig:learning_curves}, and the error rates together with model sizes and speedups are described as follows.

\noindent{\bf MNIST}.Table~\ref{tab:mnist} shows the experimental results of LBPNet on MNIST together with the baseline and previous works.
We list the classification error rates, model size, latency of the inference, and the speedup compared with the baseline CNN.
Please note the calculation of latency in cycles is made with an assumption that no SIMD parallelism and pipelining optimization is applied.
Because we need to understand the total number of computations in every network but both floating-point and binary arithmetics are involved, we cannot use FLOPs as a measure.
Therefore, we adopt typical cycle counts shown in Table~\ref{tab:cycle_count} as the measure of latencies.
For the calculation of model size, we exclude the MLP blocks and count the required memory for necessary variables to focus on the comparison between the intrinsic operations in CNNs and LBPNets, respectively the convolution and the LBP operation.

\begin{table}[!htp]
\vspace{-2mm}
\small
\centering 
\caption{The performance of LBPNet on MNIST.}
\vspace{-2mm}
\scalebox{0.9}{
\begin{tabular}{||l|rrrr||}
\hline
\multirow{2}{*}{}& \multirow{2}{*}{Error} & Size   & Latency & \multirow{2}{*}{Speedup}\\ 
                 &                        & (Bytes) & (cycles)&         \\ \hline
MLP Block       & 24.22\%                &  -      & -       & - \\ \hline
CNN (4-layer)    & 0.29\%                 &  7.00M &  3.10G  &    1X    \\
CNN (lite)       & 0.97\%                 &  1.6K &  1.843M & 1682.67X \\
BCNN             & 0.47\%                 &  1.89M &  0.306G &   10.13X\\
LBCNN            & 0.49\%                 & 12.18M &  8.776G &    0.35X\\ \hline
\multicolumn{5}{||c||}{LBPNet (this work)}\\ \hline
LBPNet (1x1)& 0.51\%                 &  1.91M & 44.851M &   69.15X\\
LBPNet (RDP)& 0.50\%                 &  1.59K &  2.609M & 1188.70X\\
\hline
\end{tabular}
}
\label{tab:mnist}
\vspace{-5mm}
\end{table}

The baseline CNN achieves the lowest classification error rate $0.29$\%, but using a significantly larger model.
The BCNN possesses a decent memory reduction and speedup while maintaining the classification.
While LBCNN claimed its saving in memory footprint, to achieve $0.49$\% error rate, $75$ layers of LBCNN basic blocks are used.
As a result, LBCNN loses memory gain and the speedup.
The $5$-layer LBPNet(1x1) with $40$ LBP kernels and $40$ $1$-by-$1$ convolutional kernels achieves $0.51$\%.
The $5$-layer LBPNet(RDP) with $39$-$40$-$80$-$160$-$320$ LBP kernels reach $0.50$\% error rate.
Although LBPNet's performance is slightly inferior, the model size of LBPNet(RDP) is reduced to $1.59$KB and the speedup is $1188.7$X faster than the baseline CNN.
Even BNN cannot be on par with such a huge memory reduction and speedup.
%We also show the performance of a reduced-size shallow CNN, CNN(lite), in the table to illustrate the benefit of LBPNets.
The worst error rate is resulted from CNN(lite).
Although we can shrink the CNN model down to the same memory size of LBPNet(RDP), the classification error of CNN(lite) is greatly sacrificed.

%\begin{wraptable}{r}{0.6\linewidth} \small
\begin{table}[!htp]
\vspace{-5mm}
\small
\centering
\caption{The performance of LBPNet on SVHN.}
\vspace{-3mm}
\scalebox{0.9}{
\begin{tabular}{||l|rrrr||}
\hline
\multirow{2}{*}{}& \multirow{2}{*}{Error} & Size   & Latency & \multirow{2}{*}{Speedup}\\ 
                 &                        & (Bytes) & (cycles)&        \\ \hline
MLP Block       & 77.78\%                &  -      & -       & - \\ \hline
CNN (10-layer)   & 6.80\%                 &  31.19M &  1.426G &  1X   \\
CNN (lite)       & 69.14\%                &  2.80K &   1.576M&  904.72X\\
BCNN             & 2.53\%                 &  1.89M &  0.312G &   4.58X\\
LBCNN            & 5.50\%                 &  6.70M &  7.098G &   0.20X\\  \hline 
\multicolumn{5}{||c||}{LBPNet (this work)}\\ \hline
LBPNet (1x1)     & 8.33\%                 &  1.51M & 91.750M &155.40X\\
LBPNet (RDP)     & 8.70\%                 &  2.79K &  4.575M &311.63X\\
\hline
\end{tabular}
}
\label{tab:svhn}
\vspace{-5mm}
%\end{wraptable}
\end{table}

\noindent{\bf SVHN}. Table~\ref{tab:svhn} shows the experimental results of LBPNet on SVHN together with the baseline and previous works. BCNN outperforms our baseline and achieves $2.53$\% with smaller memory footprint and higher speed.
The LBCNN for SVHN dataset used $40$ layers, and each layer contains only $16$ binary kernels and $512$ $1$-by-$1$ kernels.
As a result, LBCNN roughly cut the model size and the latency into a half of the LBCNN designed for MNIST.
The $5$-layer LBPNet(1x1) with $8$ LBP kernels and $32$ $1$-by-$1$ convolutional kernels achieve $8.33$\%, which is  close to our baseline CNN's $6.8$\%.
The convolution-free LBPNet(RDP) for SVHN is built with $5$ layers of LBP basic blocks, $67$-$70$-$140$-$280$-$560$, as shown in Fig.~\ref{fig:ResNet-like}.
Compared with CNN(lite)'s high error rate, the learning of LBPNet's sampling point positions is proven to be effective and economical.

%\begin{wraptable}{r}{0.61\linewidth} \small
\begin{table}[!htp]
\vspace{-4mm}
\small
\centering
\caption{The performance of LBPNet on CIFAR-10.}
\vspace{-2mm}
\scalebox{0.9}{
\begin{tabular}{||l|rrrr||}
\hline
\multirow{2}{*}{}& \multirow{2}{*}{Error} & Size   & Latency & \multirow{2}{*}{Speedup}\\ 
                 &                        & (Bytes) & (cycles)&         \\ \hline
MLP Block      & 65.91\%                &  -      & -       & - \\ \hline
CNN (10-layer)   &  8.39\%                &  31.19M &  1.426G &  1X \\
CNN (lite)       & 53.20\%                &  1.90K &   1.355M&  1052.43X\\
BCNN             & 10.15\%                &   7.19M &  4.872G & 0.29X\\ 
LBCNN            &  7.01\%                & 211.93M &193.894G & 0.01X\\ \hline
\multicolumn{5}{||c||}{LBPNet (this work)}\\ \hline
LBPNet (1x1)& 22.94\%                &  5.35M &  48.538M & 29.37X\\
LBPNet (RDP)& 25.90\%                &   1.99K &  3.265M &436.75X\\
\hline
\end{tabular}
}
\label{tab:cifar10}
\vspace{-4mm}
%\end{wraptable}
\end{table}
\noindent{\bf CIFAR-10}. Table~\ref{tab:cifar10} shows the experimental results of LBPNet on CIFAR-10 together with the baseline and previous works.
The 10-layer baseline CNN achieves $8.39$\% error rate with model size $31.19$MB.
BCNN achieved a slightly higher error rate but maintained a decent memory reduction.
Due to the relatively large number of binary kernels, the batch normalization layer in BCNN's basic blocks still needs to perform floating-point multiplications hence drags down the speed up.
LBCNN uses $50$ basic blocks to achieve the stat-of-the-art accuracy $7.01$\% error rate.
Once again, the large model of the 50-layer LBCNN with $704$ binary kernels and $384$ $1$-by-$1$ floating number kernels has no memory gain and speedup compared with the baseline.
The $5$-layer LBPNet(1x1) using $40$ LBP kernels and $40$ $1$-by-$1$ kernels achieves $22.94$\% error rate, and the $5$-layer LBPN(RDP) with $47$-$50$-$100$-$200$-$400$ LBP kernels achieves $25.90$\%.
Reducing the CNN model size to $1.9$KB, we obtain the CNN(lite) for CIFAR-10, but the performance of CNN(lite) is seriously degraded to $53.2$\%.

%This is the reason why LBCNN can learn the gradient transitions in CIFAR-10 better.

% Another reason why we are stopped at this inferior error rates is on engineering level.
% Although the inference of LBPNet is designed to be small and efficient, the backward propagation is not fully supported by CUDA BLAS and cuDNN libraries.
% The core of LBCNN is still convolution, and its forward and backward are fully supported by cuDNN library, which is highly optimized CUDA library.
% However, we build the element-wise LBP operation from scratch.
% A large portion of LBPNet's backward propagation cannot be equivalent to general matrix multiplication (GEMM) manipulation, nor accelerated by cuDNN. 
% If we overcome the GPU programming issue and improve the efficiency of backward propagation, the huge memory reduction and speedup allow us to stack a deeper LBPNet to approach the state-of-the-art performance on CIFAR-10.
\vspace{-2mm}
\subsection{Discussion}
% General discussion on LBPNet
% J: I distribute the discussion into three datasets. Do we still need this sub-section?
% J: Should we mention the CNN_light's performance on the three dataset?
\vspace{-1mm}
\begin{figure}[h]
	\vspace{-4mm}
	\centering
	\includegraphics[width=0.95\columnwidth]{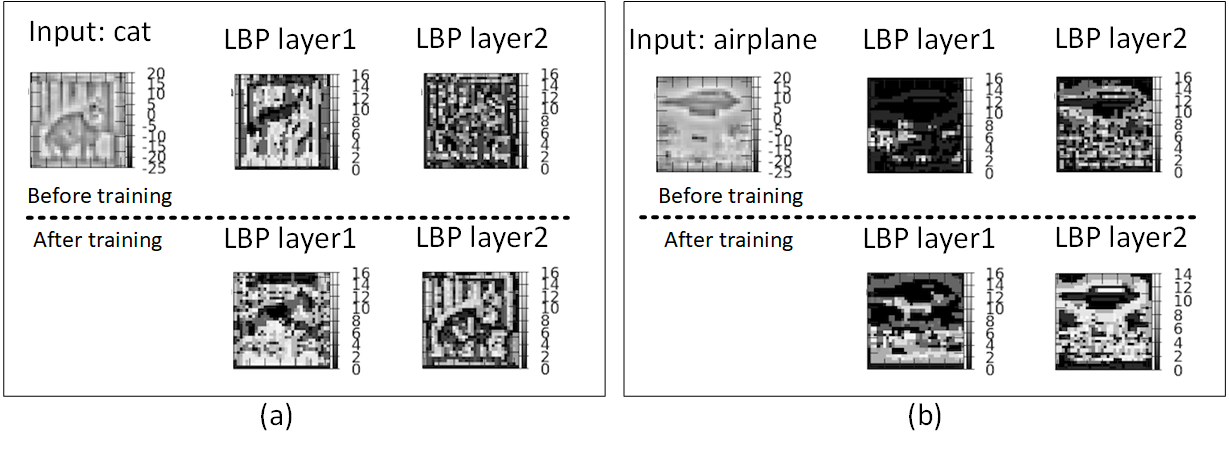}
    \vspace{-4mm}
	\caption{The transition of feature maps on the first two LBP layers.(a) an input cat image with the corresponding feature maps; (b) an input airplane image with its corresponding feature maps. The vertical bars and numbers along subfigures show the response values.}
    \label{fig:feature_maps}
    \vspace{-4mm}
\end{figure}

The learning curves of LBPNets are plotted in Fig.~\ref{fig:learning_curves}.
We also plot the baseline CNNs' error rates in blue as a reference.
Throughout the three datasets, the LBPNet(1x1)'s learning curves oscillate most because a slight shift of the local binary pattern will require the following $1$-by-$1$ convolutional layer to change a lot to accommodate the intermediate feature maps.
This is not a big issue as the trend of learning still functions correctly towards a lower error rate state.

Fig.~\ref{fig:feature_maps} shows two examples of the learning transition of feature maps on CIFAR-10.
The left hand side of Fig.~\ref{fig:feature_maps} is to learn a cat on the ground.
The right hand side of Fig.~\ref{fig:feature_maps} is to learn a airplane flying in the sky.
As we can see the transition from Epoch 1 to Epoch 300, the features become more clear and recognizable.
The cat's head and back are enhanced, and the outline of the airplane is promoted after the learning of local binary patterns.

On CIFAR-10, the main reason that stops LBPNets from getting a lower error rate is due to the discontinuity of comparison.
Unlike MNIST and SVHN having distinct strokes and outlines, CIFAR-10 is composed with daily objects, which often have gradient transitions and obscure boundaries.
LBPNets experience a hard time while extracting edges among the images in CIFAR-10, and hence the classification results are not as good as other previous works. 
LBCNN can overcome the lack of edges because it is not a genuine bit-wise operation.
LBCNN binarized and sparsified the convolutional kernels to make LBP-like, but it still took advantage of floating point arithmetic and floating point feature maps during convolving.

In summary, the learning of local binary patterns results in an unprecedentedly efficient model since, to the best of our knowledge, there is no compression/discretization of CNN can achieve the KB level model size while maintaining the an error rate as low as $0.50$\%, $8.7$\%, and $25.9$\% for MNIST, SVHN and CIFAR-10, respectively.

\vspace{-2mm}
\section{Conclusion and Future Works}
\label{sec:concl}
\vspace{-1mm}
% 1. Hardware improvement
% 2. Accuracies
% 3. Future works
%	3.1 Regularization with penalty
%	3.2 Regularization with Batch-Normalization
%\vspace{-2mm}

We have built a convolution-free, end-to-end, and bitwise LBPNet from scratch for deep learning and verified its effectiveness on MNIST, SVHN, and CIFAR-10 with orders of magnitude speedup (hundred times) in testing and model size reduction (thousand times), when compared with the baseline and the binarized CNNs. The improvement in both size and speed is achieved due to our convolution-free design with logic bitwise operations that are learned directly from scratch.
%LBPNet achieves near the state-of-the-art error rates of $0.50$\% and $8.33$\% on MNIST and SVHN, respectively.
Both the memory footprints and computation latencies of LBPNet and previous works are listed. LBPNet points to a promising direction for building new generation hardware-friendly deep learning algorithms to perform computation on the edge.
%Due to the MAC-free design, the saving of memory size is greater than $4,500$X, and the speedup is higher than $300$X compared with the baseline CNNs. The performance results are mixed but point to promising directions for future.
%The inferior performance on CIFAR-10 suggests methods to make LBPNet understand the blur outlines and gradient transitions to improve classification the error rate. %Furthermore, it appears to be a promising direction to support cuDNN in LBPNet's backward propagation. 

\vspace{2mm}
\noindent {\bf Acknowledgement} This work is supported in part by CRAFT project (DARPA Award HR0011-16-C-0037), NSF IIS-1618477, NSF IIS-1717431, and a research project sponsored by Samsung Research America. We thank Zeyu Chen for the initial help in this project.

%{\small
%\bibliographystyle{ieee}
%\bibliography{LBP_net}
%}
%%%%%%%%%%%%%%%%%%%%%%%%%%%%%%%%%%%%%%%%%%
%%%%%%%%%%%%%% Bibliography %%%%%%%%%%%%%%
%%%%%%%%%%%%%%%%%%%%%%%%%%%%%%%%%%%%%%%%%%
%\vspace{-0.08in}
%\footnotesize{
%\clearpage
%\newpage
{\small
\bibliographystyle{splncs03}
\bibliography{LBP_net}  
% \bibliographystyle{splncs}
% \bibliography{egbib}
}
\end{document}